# Compact and Wide-Force-Range Magnetorheological Hand Exoskeleton for Kinesthetic Feedback in Teleoperation

Xianlong Mai, Rui Yu, Rui Chen, *Student Member, IEEE*, Daniele Leonardis, Wenbo Li, Domenico Chiaradia, Antonio Frisoli, *Senior Member, IEEE*, Shuaishuai Sun*

***Abstract*—Hand kinesthetic feedback is crucial for restoring and enhancing human perception of force and displacement in robotic hand teleoperation. Wide perception range of force feedback is vital in rendering objects of different stiffness, especially for high-stiffness rendering. However, existing approaches struggle to achieve wide range of force feedback while maintaining compact and wearable design. This paper presents a hand exoskeleton system employing magnetorheological (MR) brakes to provide a wide range kinesthetic feedback during teleoperation. The MR brake provides a braking torque ranging from 0.01 to 0.918 N·m, while maintaining a compact structure with a total mass of only 39.8 g. The proposed exoskeleton can render fingertip forces up to 21 N while maintaining back-driving force of 1.01 N. Experimental results show that the system allows users to discriminate object stiffness using kinesthetic feedback alone. In a tele-grasping task, when both conditions achieved a 100% success rate, kinesthetic feedback led to a 28.2% reduction in object contact force compared to visual feedback only, enhancing object safety and operator immersion. Questionnaire results further indicate that kinesthetic feedback improves reliability, realism, and user confidence during teleoperation.**

## I. INTRODUCTION

Nowadays, many critical work scenarios are unsuitable or too hazardous for direct human involvement, such as mining, space exploration, explosive ordnance disposal, and operations in toxic or radioactive environments. Modern robotics is increasingly relied upon as an alternative in these situations. However, the vast uncertainties of real-world environments make it challenging for robots to achieve full autonomy [1]. At present, human-assisted teleoperation of robotic hands remains more practical and reliable. In such scenarios, hand kinesthesia plays a vital role in enabling safe and precise human robot interaction.

Kinesthetic feedback to the operator's hand is fundamental in robotic-hand teleoperation [2], providing essential information for precise and safe modulation of grasping forces and creating an immersive sense of object contact. Existing kinesthetic-feedback devices can be categorized into grounded interfaces, exoskeletons and fingertip devices. Grounded interfaces are fixed to the environment and offer accurate position tracking with a wide range and direction of force feedback [3]–[6]. However, their size and weight limit portability, making them unsuitable for mobile applications [7]. Conversely, exoskeletons and fingertip devices are portable and provide a larger workspace but must satisfy strict requirements for lightness and wearability.

Kinesthetic feedback gloves and hand exoskeletons have been extensively explored [8]–[17]. These mechanical structures maintain flexibility without restricting hand motion, though feedback mechanisms vary. For instance, Koyama et al. [9] employed a clutch mechanism with torsional springs to provide force feedback in virtual reality. While simple and effective, it offers limited bandwidth due to the spring–clutch configuration. Wu et al. [11] inserted a spring between the driving tendon and motor to deliver steady feedback with high response speed; however, its bulky structure restricts multi-finger feedback, and the force range remains limited. Ma et al. [10] introduced the RML glove featuring a compact motor–cable–linkage mechanism utilizing the worm gear's self-locking property to reach a 35 N maximum force feedback. Yet the self-locking effect yields non-adjustable force and low resolution across the dynamic range, with feedback limited to two fingers due to size constraints. The work in [17] widened the CyberGrasp glove's bandwidth to a maximum stiffness of 50 kN/m, significantly higher than its predecessor, but the rendered force (≈5 N) remained insufficient for many teleoperation tasks. Overall, each strategy presents trade-offs, with a common challenge being the realization of compact mechanisms capable of producing a wide feedback-force range.

Recently, passive brakes and clutches using MR materials show potential in balancing feedback range and size of kinesthetic device. MR materials are smart materials renowned for their tunable rheological properties in response to external magnetic fields with fast response times [18]-[19]. MR grease (MRG) is a type of MR material comprising micron-sized

This research is supported by NSFC Grant (52005474, 52105081, U21A20119), USTC start-up funding (KY2090000067), the Fundamental Research Funds for the Central Universities (WK2480000009) and Major Project of Anhui Province's Science and Technology Innovation Breakthrough Plan (202423h08050003).

Xianlong Mai, Wenbo Li and Shuaishuai Sun are with the CAS Key Laboratory of Mechanical Behavior and Design of Materials and Institute of Humanoid Robots, School of Engineering Sciences, University of Science and Technology of China, Hefei, Anhui, 230026, China.

Rui Yu is with the School of Electrical Engineering and Automation, Anhui University, Hefei, Anhui, 230601, China.

Rui Chen, Daniele Leonardis, Domenico Chiaradia and Antonio Frisoli are with the Institute of Mechanical Intelligence, School of Advanced Studies Sant'Anna (SSSA), Pisa, 56127, Italy.

*Corresponding author email: sssun@ustc.edu.cn

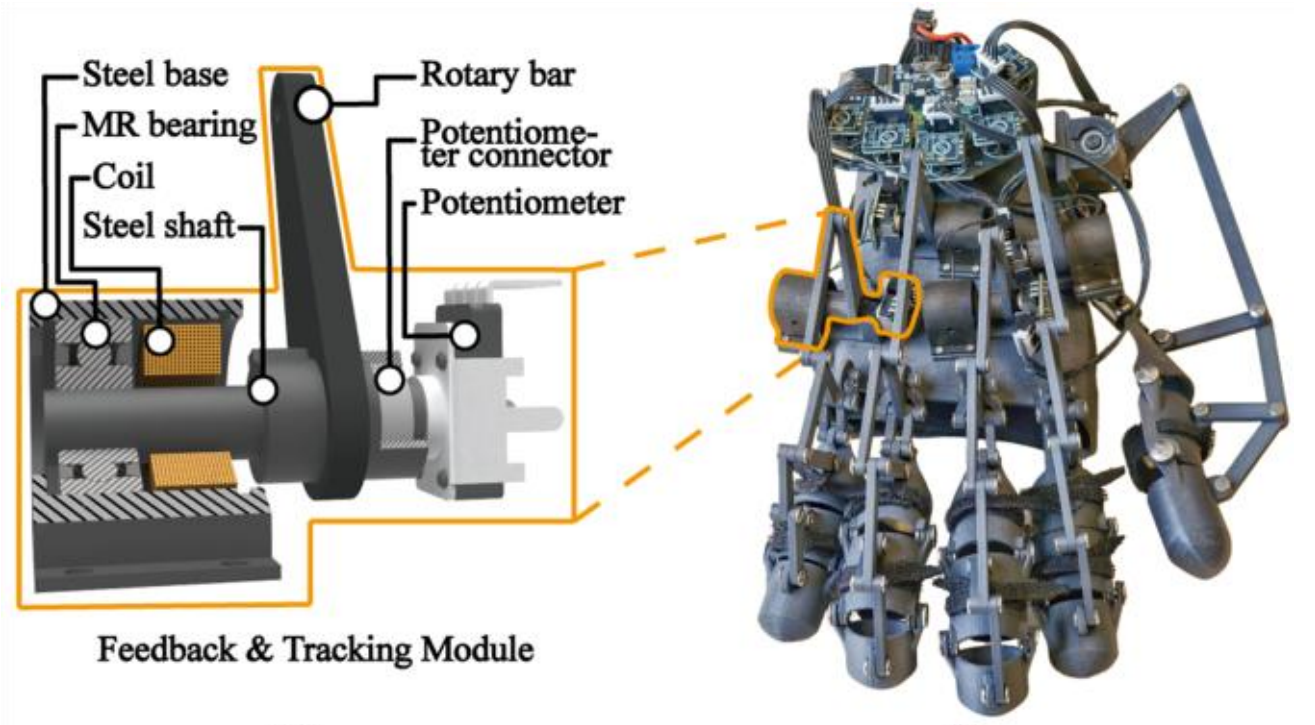

(a) (b)

Comparison of Previous Works

| Prior works | Actuator type | Actuator size / mm | Weight / g | Rendering range | Backdriving force/torque |
|---|---|---|---|---|---|
| J. Blake et al.[26] | Rotary | 21.5x36.6x28 (small) 26.5x35.5x28 (large) | 68 (small) 84 (large) | [0, 0.399] N·m (samll) [0, 0.821] N·m (large) | ≈ 0 N·m |
| Ryu S. et al.[27] | Linear | 10x10x6.5 | NaN | [1.5, 8.7] N | NaN |
| P. Kang et al.[28] | Linear | NaN | NaN | [0, 3] mm (key travel) | NaN |
| S. Choi et al.[29] | Rotary | NaN | NaN | NaN | NaN |
| Y. Kim et al.[30] | Rotary | Φ80x15 | NaN | [0.102, 0.592] N·m | 0.102 N·m |
| S. Zhu et al.[31] | Rotary | Φ13x46 | NaN | [≈ 0.1, 2.92] N·m | ≈ 0.1 N·m |
| X. Mai et al.[32] | Linear | Φ25x85 | ≈ 130 | [8.24, 1046] N | 8.24 N |
| This work | Rotary | Φ20x17 | 39.8 | [0.01, 0.918] N·m | 0.01 N·m |

(c)

**Fig.1 Prototype of the kinesthetic feedback hand exoskeleton system and comparisons.** (a) Sectional view of the feedback and tracking module. (b) Hand exoskeleton prototype. (c) Comparison of previous works.

ferromagnetic particles dispersed in grease. The high-viscosity base in MRG effectively prevents particle sedimentation while maintaining good prolonged shear stability [20]-[22], making it a preferred choice for smart rotational brake design [23] compared to MR fluid (MRF), which has higher liquidity and leakage risks in such devices [24]-[25].

J. Blake et al. [26] developed a haptic glove using an MRF-filled brake between two rotating cylinders, achieving 0.899 N·m torque at a 28:120 gear ratio. Although this torque effectively rendered stiffness, the gear transmission reduced backdrivability and increased mass. Other MR-based feedback devices include dampers [27], [28] using squeeze or shear modes to provide vertical force, and MR clutches [29] or knobs [30] operating in shear mode to deliver torque feedback. However, these designs are generally non-wearable and offer limited force range.

To address these limitations, this paper proposes an MR-bearing-based brake mechanism for kinesthetic feedback in robotic-hand teleoperation. The MR bearing is fabricated by injecting MRG into the ball gaps of a deep-groove ball bearing sealed with rubber rings. Under a magnetic field, the MRG particles align into chain structures, generating tunable rotational resistance. This configuration makes the MR bearing a compact and controllable smart component capable of delivering high torque [31]-[32]. Comparisons with previous works are summarized in Fig. 1(c). Compared with the MR brake in [26], the one proposed in this paper achieves a higher maximum torque of 0.918 N·m without gear reduction, while featuring a smaller size and lighter weight, thereby enabling full integration into a single hand exoskeleton for five-finger kinesthetic feedback. In [27] and [28], the MR actuators function as grounded, button-like haptic interfaces, which are unsuitable for wearable finger kinesthetic feedback. The actuators in [29]–[31] are considerably larger than the proposed design, making them impossible to integrate into a hand exoskeleton. As for [32], the linear actuator offers an extremely wide force range, but its size and weight are substantially greater than those of this work, and its backdriving force reduces transparency during haptic operation. Consequently, the proposed MR brake achieves a smaller size, lighter weight, and larger rendering range among the compared works.

The main contributions of this paper are summarized as follows:

1. A compact hand exoskeleton integrating MR brakes is proposed, which offers a highly integrated, lightweight solution with a wide force feedback range for kinesthetic feedback in teleoperation.
2. Theoretical models and simulations of the hand exoskeleton and MR bearing are established to predict the fingertip force feedback and theoretical braking torque.
3. System characterizations are conducted to validate the theoretical models and demonstrate the performance of the proposed design.
4. User experiments are conducted and showed improvements in telepresence, as well as in the confidence and safety of telemanipulation.

The article is structured as follows: Section II describes mechanical and electrical design of the MR brake, hand exoskeleton, and robotic hand. Section III discusses the kinematic modeling of the hand exoskeleton and MR bearing. Section IV and V reports system characterization and user studies. Section VI provides discussion and conclusions.

## II. Mechanical and Electrical Designs

### A. Mechanical Design of Kinesthetic Feedback System

The MR brake provides a wide range of force feedback to each finger during robotic hand teleoperation. As shown in Fig. 1(a), the feedback and tracking module includes: MR brake (consists of a steel base, MR bearing, coil and steel shaft), rotary bar, potentiometer connector, and potentiometer. The outer and inner rings of the MR bearing are interference-fitted to the steel base and steel shaft, respectively. Since the steel base is fixed to the dorsal surface of the hand exoskeleton (as shown in Fig. 1(b)), the MR bearing effectively controls the rotary resistance of the steel shaft. A coil positioned adjacent to the bearing generates a magnetic field that adjusts the desired braking torque, thereby modulating the force feedback applied to the finger. This coil is clearance-fitted with both the base and shaft, and remains stationary relative to the base to prevent short circuits during shaft movement. To optimize magnetic flux concentration, both the steel base and the shaft are fabricated from low-carbon steel (ASTM 1020). A potentiometer, connected to the shaft via a customized 3D-printed connector (Polylactic Acid, PLA, Bambu Lab Co., Ltd.), synchronously tracks finger motion during kinesthetic feedback. The rotary bar, 3D-printed with PLA and slot-connected to the steel shaft, simultaneously drives the potentiometer for motion tracking and transmits the force feedback from the MR brake. The detailed design specifications and materials can be found in Fig. S1 in Supplementary Material.

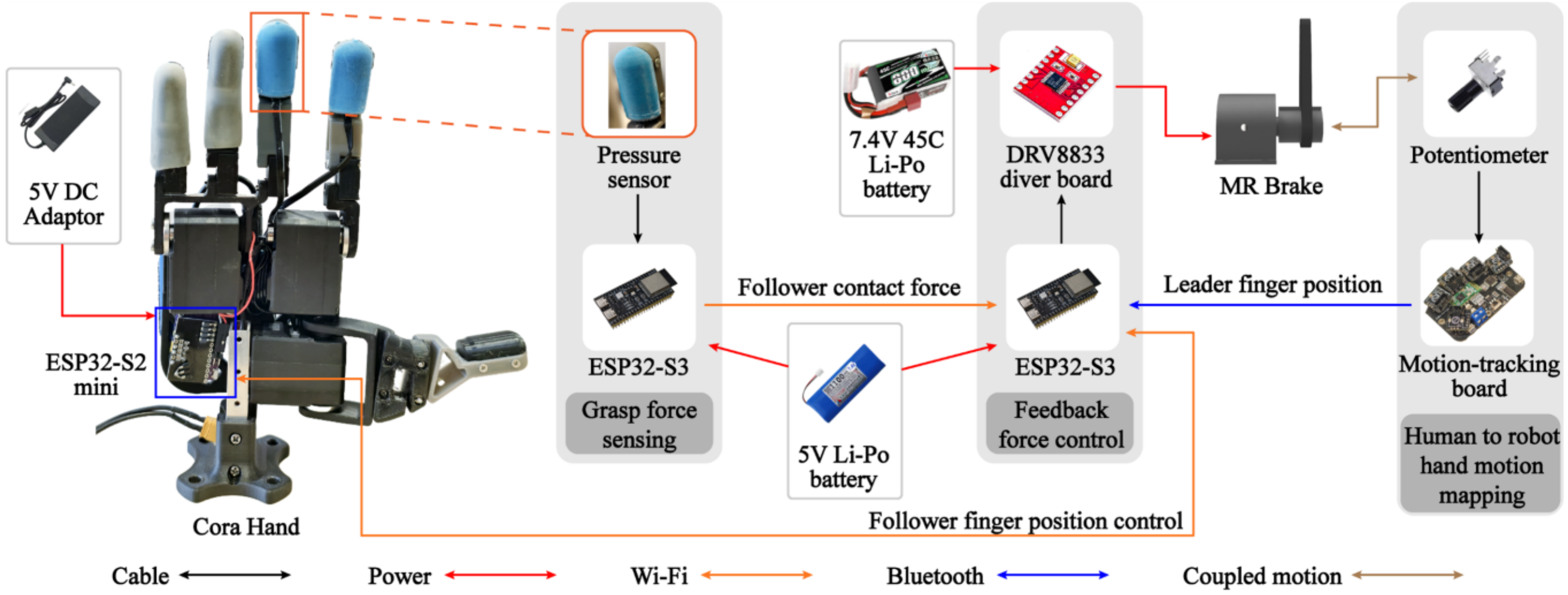

**Fig.2 Electrical architecture and data communication scheme for kinesthetic feedback-assisted teleoperation.**

The hand exoskeleton has five degrees of freedom and adopts linkage structure for force and motion transmission. Velcro straps secure the remaining phalanges, providing adjustable sizing to accommodate varying finger diameters. All components are 3D-printed using carbon-fiber-reinforced PLA (PLA-CF, Bambu Lab Co., Ltd.), and aluminum alloy rivets are used to connect the joints. This construction ensures a lightweight structure that can be rapidly customized for different hand sizes. Additionally, the electrical circuit for finger motion tracking is integrated into the dorsal side of the exoskeleton, improving overall system compactness. The complete system, as shown in Fig. 1(b), weighs 495 g, representing a 22.7% reduction compared to the design in [26] while maintaining kinesthetic feedback capabilities for all fingers.

### *B. Teleoperated Robotic Hand*

To demonstrate the proposed kinesthetic feedback system in a cost-effective and accessible manner, we utilized a custom-built 3-degree-of-freedom robotic hand named Cora Hand, based on the prototype described in [33] (as shown in Fig. 2). The thumb abduction and index finger flexion are independently actuated by separate servo motors, while the flexion of the remaining three fingers is mechanically coupled and actuated by a single motor. To make an intuitive and simple mapping between the leader-follower system, in this paper, the movements of thumb and index finger in the exoskeleton side are mapped to the corresponding fingers' movements of the Cora Hand, and the movement of middle exoskeleton finger is mapped to the coupled fingers' (middle, ring and little fingers) movement of the Cora Hand. Feetech ST3215 servomotors are implemented and operated with a proprietary position control loop, with modulated maximum current saturation. To measure fingertip forces during grasping, two sensors are mounted on the finger pads of the index and middle fingers. Each sensor consists of a pressure sensor (Bosch BMP280) embedded in an airtight silicone rubber dome, enabling the detection of fine contact transients with a relative accuracy of 0.1 mbar over the whole contact area of the finger pad.

### *C. Electrical Components and communication protocols*

The electrical components and data communication designs employ closed-loop control strategies, as illustrated in Fig. 2 (right). Initially, the hand motions of the exoskeleton wearer (referred to as the "leader") are mapped to those of the robotic hand ("follower"). Since both systems feature single-degree-of-freedom fingers, the angular range of the robotic finger is scaled to the angular range of the hand exoskeleton, which is measured by potentiometer. Fingertip contact forces on the follower side are detected by the pressure sensors. The signals are digitized using the ADC converter on an ESP32-S3 microcontroller board, and transmitted via Wi-Fi to another ESP32-S3 board on the leader side. The microcontroller then adjusts the MR brake current through a driver board (DRV8833) using Pulse Width Modulation (PWM) signals. The MR brake's rotary motion is mechanically coupled with the potentiometer rotation. Potentiometer data are processed by a motion-tracking board based on an STM32 microcontroller, and subsequently sent via Wi-Fi UDP to the control board (ESP32-S2) embedded on the Cora hand. Here the control loop is processed at 100 Hz, performing reading of the finger angular position references from the glove, remapping to the robotic fingers, and driving of the servomotors. In this setup, kinesthetic feedback is directly coupled with the motion of the teleoperated robotic fingers.

## III. Theoretical Analysis

### *A. Kinetic model of the exoskeleton*

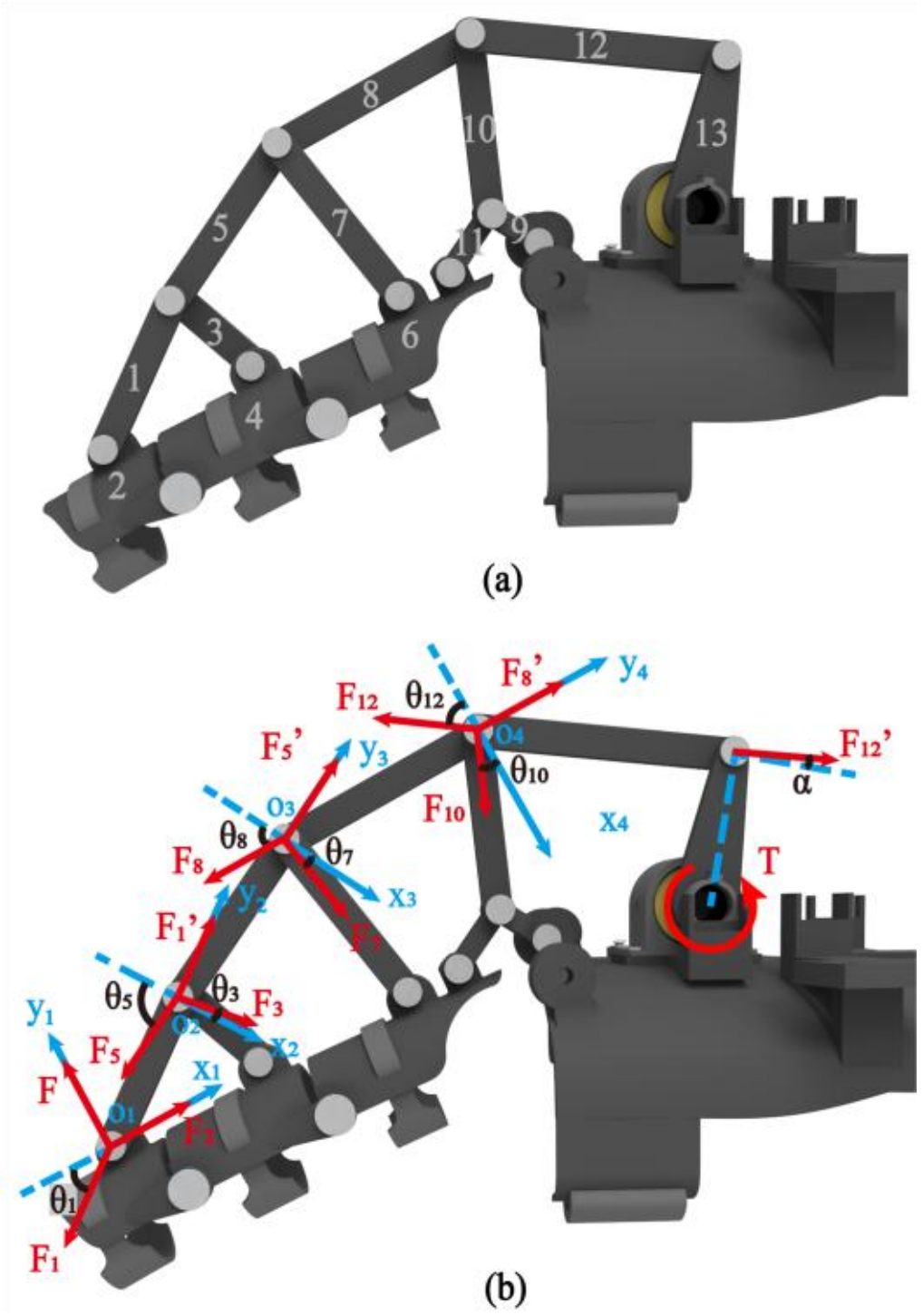

**Fig. 3 Kinetic model of the hand exoskeleton.** (a) Number of the two-force members in the kinetic model. (b) Definition of Cartesian coordinates, forces and angles related to the kinetic model.

To build the relationship between the output torque of the MR brake, *T*, and the force feedback on the fingertip, *F*, a kinetic model of the hand exoskeleton is built, as shown in Fig. 3(a) and Fig.3(b). The model is simplified as a rigid body model in the plane of middle finger and takes the exoskeletons and linkages as two-force members. The number of each two-force member is indicated in Fig. 3(a). In Fig.3(b), four sets of Cartesian coordinates are built in blue and named $x_1o_1y_1$, $x_2o_2y_2$, $x_3o_3y_3$ and $x_4o_4y_4$ respectively. The fingertip force *F* is perpendicular to member number 2. Axis $y_1$ is collinear to force *F*, and axes $y_2$, $y_3$ and $y_4$ are collinear with the members 1, 5 and 8, respectively. Axes $x_1$, $x_2$, $x_3$ and $x_4$ are perpendicular to $y_1$, $y_2$, $y_3$ and $y_4$, respectively. The origins of the four coordinates are fixed at the hinges which join members 1 and 2, members 1, 3 and 5, members 5, 7 and 8, and members 8, 10 and 12. All the coordinates are right-handed systems. The axis forces of the members are marked as $F_i$ and $F_i{}'$. The index number $i$ stands for its corresponding member's number. $F_i$ and $F_i{}'$ are equal in size and opposite in direction. The output torque of the MR brake is marked as *T*, which is applied on one end of member 13. The parameters $\theta_1$, $\theta_3$, $\theta_5$, $\theta_7$, $\theta_8$, $\theta_{10}$, $\theta_{12}$, are the angles between $F_1$ and $x_1$, $F_3$ and $x_2$, $F_5$ and $x_2$, $F_7$ and $x_3$, $F_8$ and $x_3$, $F_{10}$ and $x_4$, $F_{12}$ and $x_4$, respectively. Angle $\alpha$ is the angle between $F_{12}{}'$ and the perpendicular direction of member 13. In the coordinate $x_1o_1y_1$, the governing equations according to equilibrium of forces can be obtained:

$$F_1 \times \sin\theta_1 = F \tag{1}$$

$$F_3 \times \cos\theta_1 = F_2 \tag{2}$$

From (1), we have

$$F_1 = \frac{F}{\sin\theta_1} \tag{3}$$

Similarly, in coordinate $x_2o_2y_2$, we have the following equations according to equilibrium of forces:

$$F_5 \times \sin\theta_5 + F_3 \times \sin\theta_3 = F_1 \tag{4}$$

$$F_5 \times \cos\theta_5 = F_3 \times \cos\theta_3 \tag{5}$$

By substituting (3) and (4) into (5), we have

$$F_5 = \frac{F \times \cos\theta_3}{\sin\theta_1 \times \sin(\theta_5 + \theta_3)} \tag{6}$$

In the same way, by using force equilibrium method in coordinates $x_3o_3y_3$ and $x_4o_4y_4$ respectively, we finally have the relationship between the fingertip force *F* and the output torque *T*:

$$F = \frac{T}{l_{13}} \cdot \frac{\sin\theta_1 \sin(\theta_5 + \theta_3) \sin(\theta_8 + \theta_7) \sin(\theta_{12} + \theta_{10})}{\cos\theta_3 \cos\theta_7 \cos\theta_{10} \cos\alpha} \tag{7}$$

where $l_{13}$ is central length between the two hinges of member 13.

*B. Model of the MR bearing*

The behaviour of MRF is usually described by the Bingham plastic model:

$$\tau = \tau_{yd}(B) + \eta\dot{\gamma} \tag{8}$$

where $\tau$ is shear stress, $\dot{\gamma}$ is shear rate, $\eta$ is viscosity, $B$ is the applied magnetic flux density, and $\tau_{yd}(B)$ is the dynamic yield stress of the MRF [34]. While the MRF and MRG have different viscosities, their dispersion and flow characteristics are quite similar, which means the same governing rule can be used in the analysis of MRG. In this work, when a magnetic field is applied to the MR bearing, the braking torque can be mainly originated from two parts: the yield stress and shear stress of the MRG. S. H. Mousavi and H. Sayyaadi proposed a theory to calculate these two parts based on the slab method in [35]. In this theory, the torque calculation has been simplified and focused on calculation of single steel ball in the bearing. The yield stress in the single-steel-ball model can be described as the drawing force $F_d$ using the slab method, and the shear stress includes Coulomb shear stress $F_\tau$ and viscous shear stress $F_\eta$. The total resistance force $F_r$ of the single-steel-ball model fulfills the following relationship:

$$F_r = F_d + F_\eta + F_\tau \tag{9}$$

Detailed calculation of $F_d$, $F_\eta$ and $F_\tau$ can be found in the Supplementary Method in the Supplementary Material. With the total resistance forces of the single-steel-ball model, the total torque generated by the MR bearing can be calculated as:

$$T' = N_r\Gamma_r \int_{-\psi}^{\psi} r_r F_r [R_c + (1 - \cos\beta)(r_r + h_{min})] d\beta \tag{10}$$

where $N_r$ is the number of steel balls, $\Gamma_r$ is the effective coefficient, $R_c$ is the distance between bearing center and the raceway, $\psi$ is the effective magnetic field angle in the axis plane. The torque $T'$ is equal to the torque T mentioned in (7).

*C. Magnetic field simulation and theoretical braking torque calculation*

To evaluate the spatial distribution of the magnetic field in the MR brake, the magnetic flux density – magnetic field strength (*B-H*) curve of the MRG was obtained by testing MRG samples on a Vibrating Sample Magnetometer (SVM), as shown in Fig. 4(a). With the data of the MRG, a magnetic field simulation was conducted using COMSOL Multiphysics on the MR brake. Fig. 4(b) shows the magnetic field distribution in a sectional view of the MR brake (I=1.5A). The results show that most of the magnetic flux forms a closed loop through the steel base, MR bearing, and steel shaft. The use of low-carbon steel for both the base and shaft provides high magnetic permeability and effective paths for flux concentration. As a result, the majority of the magnetic flux passes through the gap between the steel balls and the outer ring. This distribution induces the formation of chain-like structures by the ferromagnetic particles in the MRG, providing resistance to bearing rotation. This resistance is tunable and depends on the local magnetic flux density within the gap, as described in the analysis of MR bearing model. To calculate the theoretical braking torque $T'$, a probe was placed in the gap (as shown in Fig. 4(b)) in the COMSOL simulation to measure the magnetic flux density norm under different current levels. As depicted in Fig. 4(c), the magnetic flux density shows a linear correlation with the applied current. A linear fitting was applied to the probe values, and the *B-I* relation can be described as:

$$B = 0.249I + 0.566 \text{ (T)} \tag{11}$$

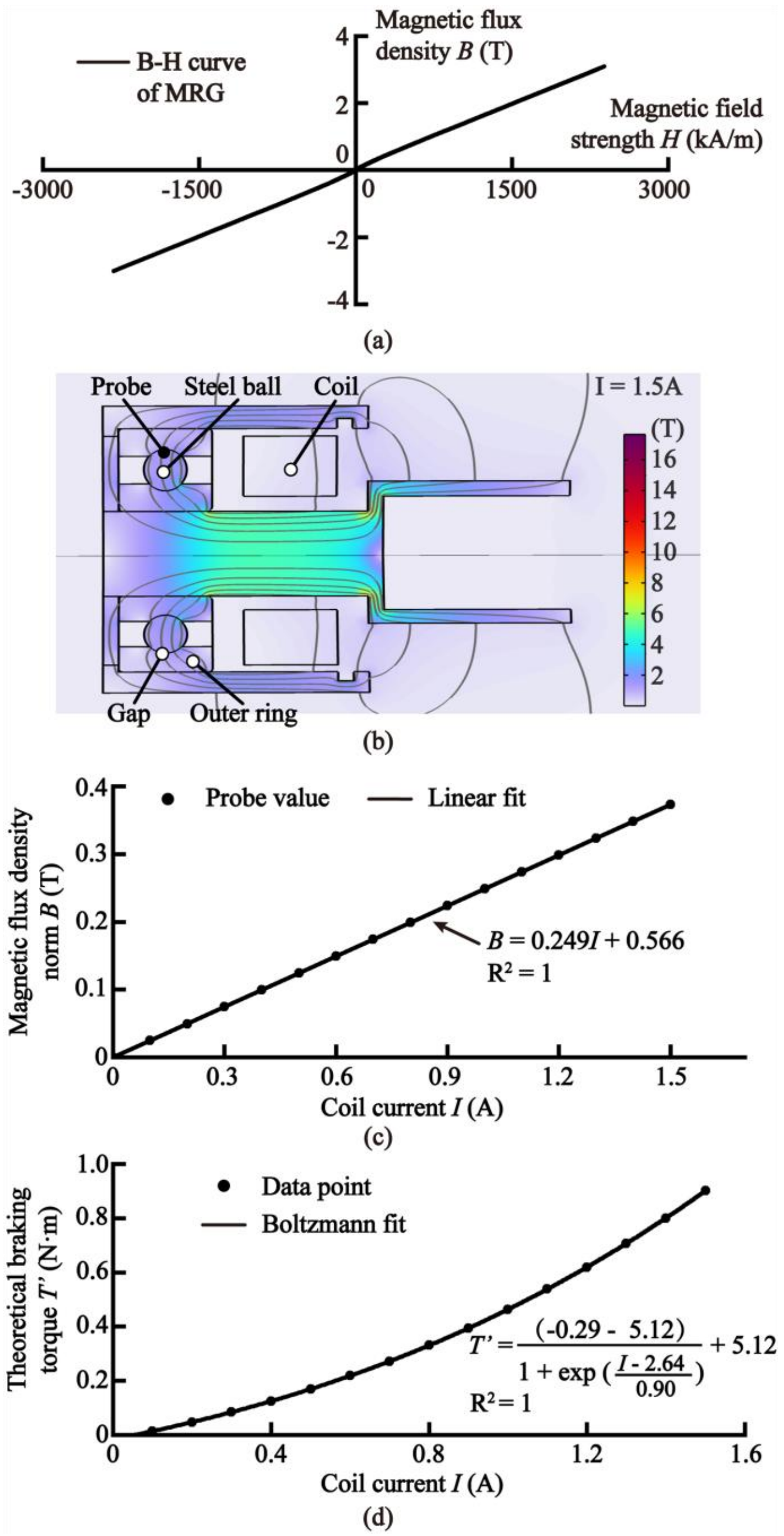

**Fig. 4 Magnetic field simulation and theoretical braking torque evaluation of MR brake.** (a) The magnetic flux density –magnetic field strength (*B-H*) relation of MRG. (b) Magnetic field simulation result of the MR brake when current intensity is 1.5 A. (c) Magnetic flux density norm at the probe shown in (b). (d) Results of theoretical braking torque calculation.

The viscosity and yield stress of the MRG were measured using a rotational rheometer, and the relation between yield stress $\tau_y(B)$ and magnetic flux density $B$ can be described as:

$$\tau_y(B) = \frac{(-7.378 - 131.858)}{1 + e^{B-0.659}} + 131.858 \tag{12}$$

By first substituting (11) and (12) into the equations in the Supplementary Method, and then substituting the resulting expressions into (9), the theoretical braking torque $T'$ is obtained and depicted in Fig. 4(d), where the data points are fitted by a Boltzmann function.

## IV. System Characterization

### *A. Response and braking torque evaluations of MR brake*

The braking torque generated by the MR brake plays a crucial role in delivering effective kinesthetic feedback. A wide tunable range of output torque as well as rapid response, are essential characteristics for kinesthetic rendering. To evaluate these characteristics, an experimental setup was constructed to measure the braking torque of the MR brake under varying current levels.

As shown in Fig. 5(a), the MR brake was driven by an AC motor (SVGA-15H11AD, TODE Technologies Co., Ltd.), with a torque sensor (TFS-T104A-30N·m, Dongguan Tefeisi Technology Co., Ltd., China) connected via couplings to record the resistance torque.

The response test was first conducted. The motor was first set to a constant rotational speed of 8 rpm (revolutions per minute). Subsequently, a step voltage of 1.3 V (current of 1 A) was applied to the MR brake coil, generating a magnetic field that makes the MR brake generate braking torque. The torque sensor recorded the resistive torque generated by the MR brake at a sampling rate of 5 kHz. The test was repeated five times, with 10 seconds of bidirectional rotation between trials to reconfigure the MRG. After data interception, low-pass filtering, denoising, and smoothing, the processed results are presented in Fig. 5(b). MR brakes are typically modeled as first-order systems, in which the response time is defined as the rise time ($t_{rise}$)) between 0% and 63.2% of the steady state torque $T_s$. As shown in Fig. 5(b), the average response time is 33.26 ± 4.31 ms, indicating that the MR brake achieves a response level comparable to most existing MR brakes and dampers. The coil voltage curve in Fig. 5(b) (details can be found in Fig. S2 in Supplementary Material) show the average response time (from zero to 90% of the target voltage) of coil driver is 13.5 ms.

Following the response test, experiments were performed to measure the maximum braking torque of the MR brake. In this experiment, the brake operated as a passive load with varying coil current from 0 A to 1.5 A in 0.25 A increments. For each trial, the desired current level was applied before motor was driven at a speed of 8 rpm. Each current condition was tested five times, with the motor rotating through a fixed angular displacement of 120° in the same direction. Between tests, the brake was demagnetized (details can be found in Section IV.D) and bidirectionally rotated for 10 seconds to reconfigure the MRG and eliminate magnetic hysteresis. The mean maximum braking torque values are shown in Fig. 5(c), with a Boltzmann function fitted to all data points. Overall, the measured braking torque exhibits a similar trend to both the experimental yield stress $\tau_y(B)$ of the MRG and the theoretical torque $T'$ presented in Fig. 4(d). As shown in Fig. 6, the experimental values correspond well to the theoretical values when current levels are equal to and below 0.75 A. At 0 A, a low resistance torque of 0.01 N·m is observed and mainly attributed to the MRG's inherent viscosity and its interaction with the bearing. As the current increases to 1.5 A, the torque reaches 0.918 N·m, representing the maximum braking torque within the coil's current-carrying limit. Compared with the results reported in [26], the torque output of the proposed MR brake at 1 A is 3.19 times and 6.59 times higher than those of the large and small MR brakes (without gear reduction) reported in [26], respectively.

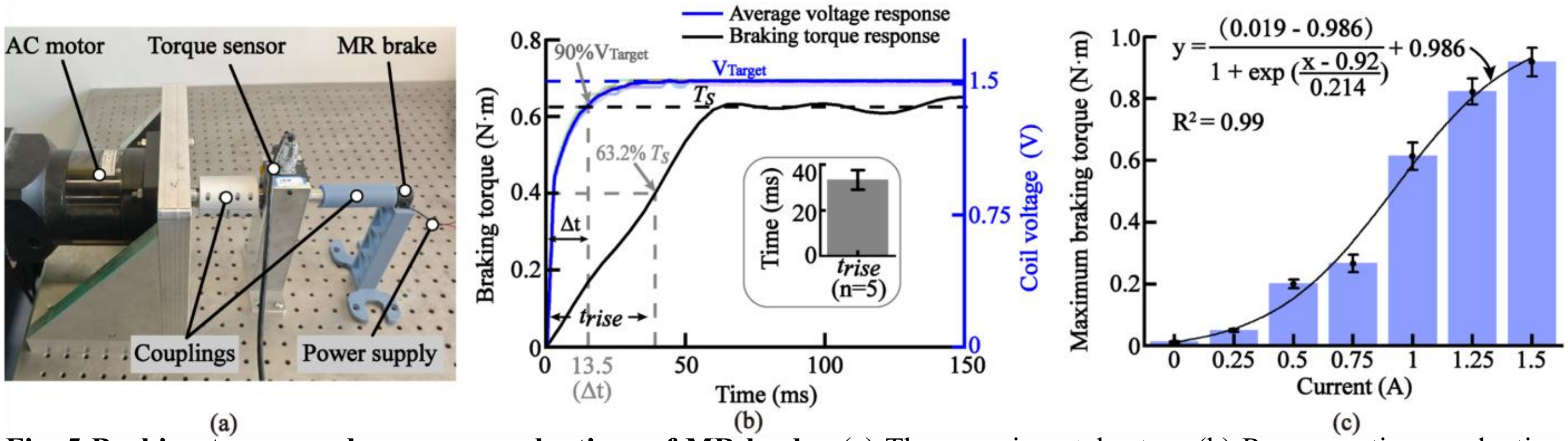

**Fig. 5 Braking torque and response evaluations of MR brake.** (a) The experimental setup. (b) Response time evaluation results. (c) The maximum braking torques at variable current intensities. Error bars represent ±1 SD about the mean (n=5).

### B. Force feedback evaluation

To quantify the magnitude of force feedback delivered to the finger, a load cell was mounted on the hand exoskeleton beneath the middle fingertip, as shown in Fig. 7(a). The coil current was varied from 0 A to 1.25 A in 0.25 A increments. For each current level, five trials were performed, with 20 s of rest and bidirectional MR brake rotation between trials to eliminate magnetic hysteresis. One participant (male, 27 years old) was instructed to flex his middle finger from a fully extended position. The same finger posture was maintained for all trials to minimize measurement bias associated with different joint angles. The angular parameters used in the kinematic model were measured and are presented in Table I. Theoretical force feedback was then calculated by substituting the angles from Table I and braking torques from Fig. 5(c) into Eq. (7). As shown in Fig. 7(b), the experimental and theoretical force feedback are plotted together for direct comparison. When the applied current was low (below 1 A), the fingertip force feedback was easily overcome, and the mean values of the flexion forces were taken as the force feedback at these current levels. For higher currents (1 A and 1.25 A), the participant was required to exert greater flexion effort until the force feedback was overcome; in these cases, the recorded values represent the maximum achievable force feedback on the middle finger.

At 1.25 A, the experimental force feedback reached 21 N, which was nearly the maximum force the participant could overcome; thus, the experimental current range was limited to 1.25 A. To evaluate the accuracy of the kinematic model, the relative error was computed as the absolute difference between theoretical and experimental values divided by the theoretical value, as illustrated in Fig. 7(c). At zero current input, the relative error is notably higher than at other current levels, primarily because frictional effects constitute a larger proportion of the small force feedback. For currents between 0.25 A and 1.25 A, the relative errors remain below 10%, and particularly below 5% for currents from 0.5 A to 1.25 A, confirming the general accuracy of the kinematic model.

TABLE I

EXPERIMENTALLY MEASURED KINETIC MODEL PARAMETERS

| | | | |
|---|---|---|---|
| $\theta_1$ | $\theta_3$ | $\theta_5$ | $\theta_7$ |
| 44° | 35° | 53° | 29° |
| $\theta_8$ | $\theta_{10}$ | $\theta_{12}$ | $\alpha$ |
| 63° | 23° | 52° | 3° |

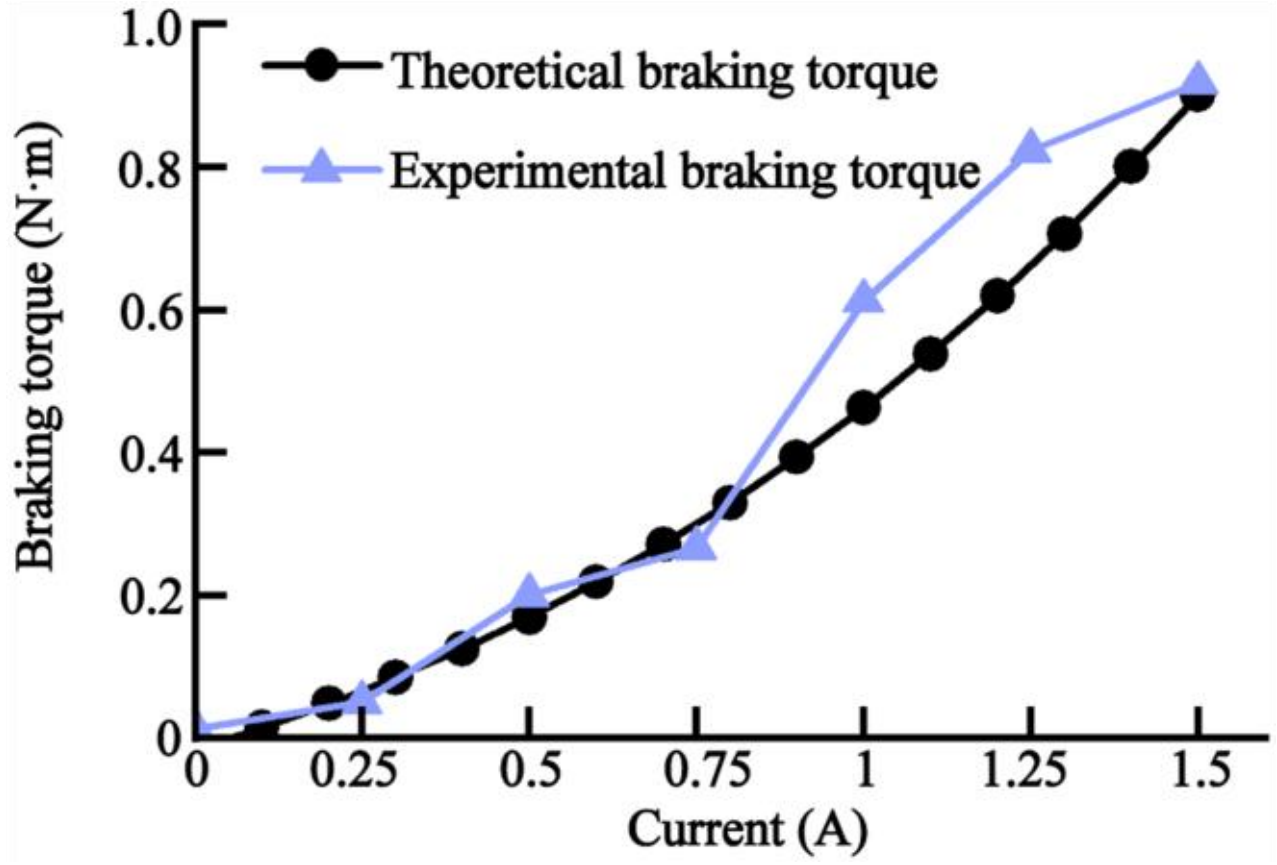

**Fig. 6 Comparison between theoretical and experimental braking torques.**

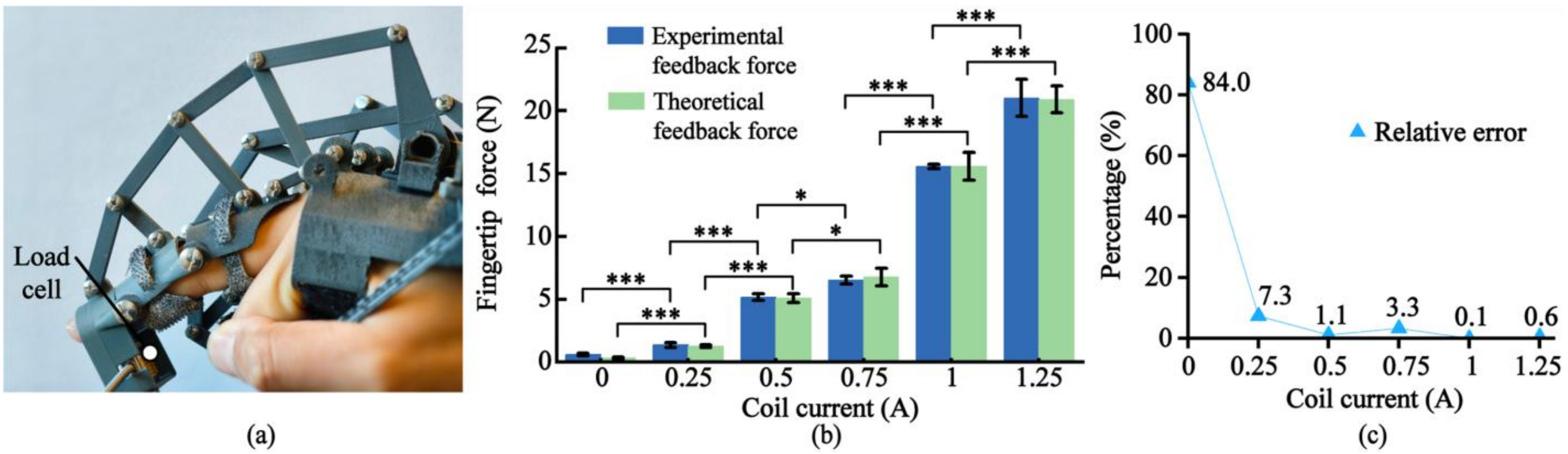

**Fig. 7 Fingertip force feedback evaluation.** (a) Schematic diagram of measuring force feedback on middle fingertip. (b) Measured fingertip forces at different coil current intensities. (c) Relative errors between experimental and theoretical force feedback. Error bars represent ±1 SD about the mean (n=5). Asterisks denote statistical significance (Paired sample t-test, *P < 0.05, **P < 0.01, ***P < 0.001).

## V. User Studies

All experiments were carried out in accordance with the declaration of Helsinki on research involving human subjects, and were approved by The First Affiliated Hospital of USTC (NO. 2023KY239). All subjects provided explicit written consent to participate in these studies.

In sections V.A and V.C, twelve participants (four females and eight males, aged 22–34 years (mean=26.7, SD=3.5)) were recruited. None reported any known sensorimotor impairments. In Section V.B, one participant (male, 27 years old) was involved and not reported any known sensorimotor impairments. Detailed information of the participants can be found in Table S1 in Supplementary Material.

### *A. Objects classification*

As shown in Fig. 8(a), participants were instructed to teleoperate the robotic hand mounted on a UR-5 robotic arm and to classify objects of different stiffness using only kinesthetic feedback from the hand exoskeleton via the index finger. Four object types were used (all of them are 5x5x5 cm in size): Air (no object or contact force), 15D sponge cube (about 0.25 N/mm), 35D expandable polyethylene (EPE) cube (about 15 N/mm), and PLA 3D-printed cube (50% filling rate, >100 N/mm). To block visual and auditory feedback, participants wore eye masks and noise-isolating earmuffs, as shown in Fig. 8(b). They sat approximately 2 meters from the robotic hand. Prior to the formal experiment, each participant received 5 minutes of training to familiarize themselves with the kinesthetic sensation associated with each object. Force feedback was generated using a calibration method matching the contact pressure measured by fingertip sensors to a contact force. The relationship between sensor pressure output and contact force was calibrated using a load cell, and the linear fitting result is shown in Fig. 8(c). As the output exhibited an approximately linear relationship with contact force, a linear mapping was adopted to convert contact pressure into contact force intensity. During the experimental trials, participants were asked to determine the presentation order of the four objects by bending the index finger to touch them in a randomized sequence. Each participant completed three trials.

The results are summarized as a confusion matrix in Fig. 8(d). High classification accuracy was observed for the two extreme cases (i.e., air and PLA cube). For intermediate objects (sponge and EPE), the accuracies were 61.1% and 72.2%, respectively. These results demonstrate the system's effectiveness in rendering kinesthetic feedback for objects with large differences in stiffness, while also highlighting limitations in distinguishing objects with subtle stiffness variations.

### *B. Leader-follower interaction data analysis*

To evaluate the dynamic relationship between finger force and position, a teleoperated grasping experiment was conducted using the proposed leader–follower system. A rigid, PLA 3D-printed cylinder (diameter of 89 mm, height of 150 mm) was used as the grasping object. The operator was asked to grasp the object using index finger and the thumb, hold the grasping for a while, and then release it. The forces at the robotic and operator's index fingers, as well as the potentiometer angle (operator's finger position), were measured.

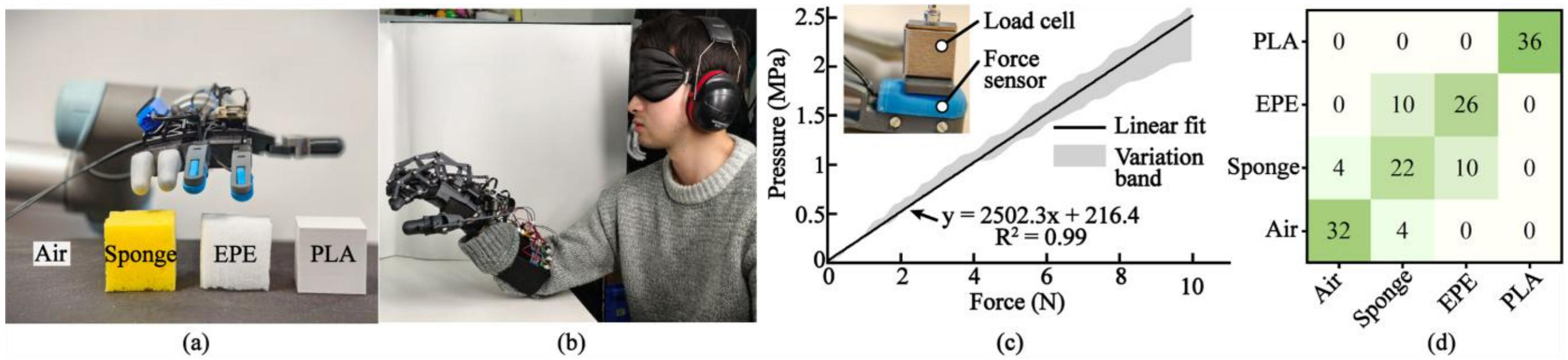

**Fig. 8 Experimental setup and results of objects classification.** (a) Experimental setup and four types of objects. (b) Participant situation in the experiment. (c) Calibration pressure sensor value and applied force. (d) Experimental result in the form of confusion matrix.

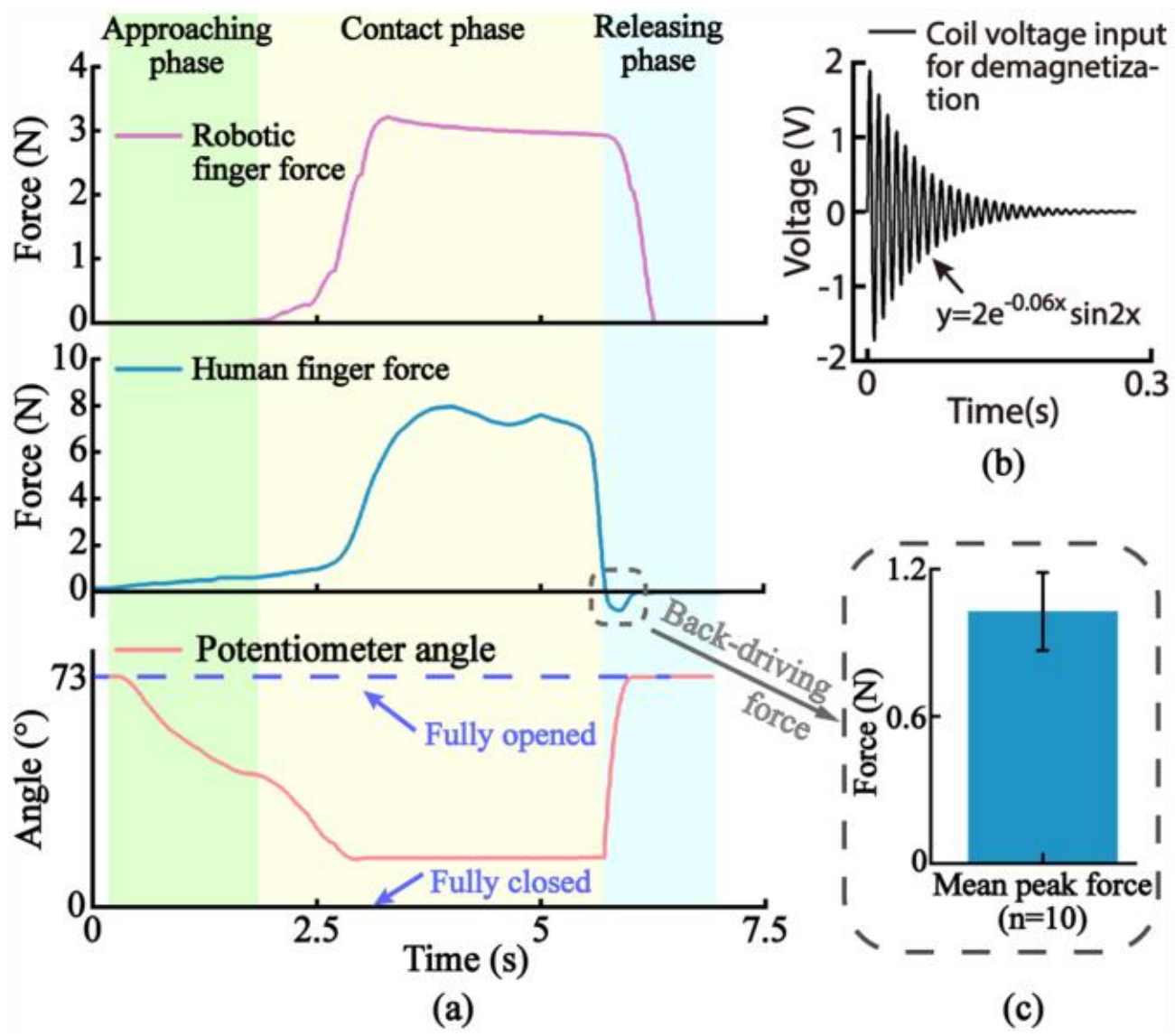

**Fig. 9 Leader-follower interaction data analysis in tele-grasping. (a)** Contact force on robotic hand, index fingertip force and potentiometer angle in tele-grasping. (b) The coil voltage input (an attenuated sine wave) for eliminating magnetic hysteresis of the MRG bearing (demagnetization). (c) Mean back-driving force on leader fingertip. Error bars represent ±1 SD about the mean (n=10).

As shown in Fig. 9(a), the grasping process is divided into three phases: approaching, contact, and releasing. In the approaching phase, the robotic hand was not yet in contact with the object, so the robotic finger contact force was zero. However, due to the inherent friction from the MR brake and mechanical structure, the user's index fingertip force increased slightly, remaining below 1 N. During the contact phase, the robotic finger force increased sharply upon contact. As the MR brake generated torque feedback and the force feedback on wearer's fingertip rose rapidly, the wearer perceived resistance and stopped applying force. This behavior is reflected in the plateau of the finger position curve. The force feedback was observed a little delayed due to communication delay and the compliance of mechanical structures. When the robotic hand maintained its grasp, both force and position readings stabilized, though minor fluctuations in force feedback were observed due to slight hand movements. When the wearer attempted to release the object, the coil in the MR brake was given an attenuated sine voltage (shown in Fig. 9(b)) to eliminate the magnetic hysteresis (demagnetization). The shape of the voltage curve was chosen based on experience. In the receding phase, the human finger force decreased to zero first, followed by a slight negative force resulted from reverse finger motion. This force is the back-driving force that prevents finger movement and should be as low as possible for kinesthetic feedback device. Repeated experiments were conducted to quantify the maximum back-driving force after demagnetization, and the mean peak force is plotted in Fig. 9(c). On average, the maximum back-driving force is 1.02 N, which shows both the effectiveness of demagnetization and negligible restriction on wearer's finger movement.

To further evaluate the influence of demagnetization on eliminating magnetic hysteresis, the output torque of the MR brake was measured under four conditions: 1) coil powered on; 2) coil powered on and then powered off; 3) coil powered on, powered off, and then the MR brake manually rotated; 4) coil powered on, powered off, and then the MR brake demagnetized. The MR brake was fixed to a stationary base. A dynamic torque sensor (TEFEISI SENSOR TECHNOLOGY, 10 N·m) was connected between the output shaft of the MR brake and a motor, which drove the brake with a sinusoidal motion within ±45°. As shown in Fig. S3 in the Supplementary Material, condition one exhibited the highest peak torque. The torque reduction between consecutive peaks was caused by the fracture and restructuring of magnetic particle chains under the applied external torque. In condition two, the peak torque decreased but remained within ±0.2 N·m. This residual torque was attributed to magnetic hysteresis in the MR bearing and would produce an unrealistic sensation for the user. In condition three, manually rotating the brake five times over [−90°, 90°] effectively eliminated the residual particle chains resulting from magnetic hysteresis, yielding a much lower peak torque (within ±0.05 N·m). Condition four initially followed the same procedure as condition two, but subsequently a demagnetizing voltage (Fig. 9(b)) was applied. The corresponding result in Fig. S3 (purple line) showed a significant reduction in peak torque compared with condition two. Compared with condition three, the demagnetization method achieved a similar torque level at most positions. A few small torque peaks (all within 0.15 N·m) were observed at certain positions and were eliminated by the subsequent motion.

### *C. Evaluation on teleoperating task*

To further investigate the impact of kinesthetic feedback, a more complex teleoperation task was designed for assessment. As shown in Fig. 10(a), the robotic hand was mounted on a UR-5 robotic arm. Participants sat 3 meters away from the robotic hand and received visual feedback through a camera streaming. Each trial consisted of three steps: grasping a cylindrical object using the index finger and thumb, pressing a key to lift and lower the object, and finally releasing it. Participants were asked to determine the grasping force they use based on kinesthetic feedback and/or visual feedback, to grasp the object securely without using too much force. An additional button-pressing step was required before releasing the object to demagnetize the MR brake and eliminate remanence when haptic feedback was enabled. Each completed 10 trials under two feedback conditions: V with visual feedback only, and HV with both visual and haptic feedback enabled. Between trials, a 10-second rest was provided, during which the object was repositioned to its initial location.

The robotic finger force was recorded throughout each trial, and the mean values for both conditions are shown in Fig. 10(b). Data were analyzed using a paired-sample t-test. There was a statistically significant difference between Condition HV and

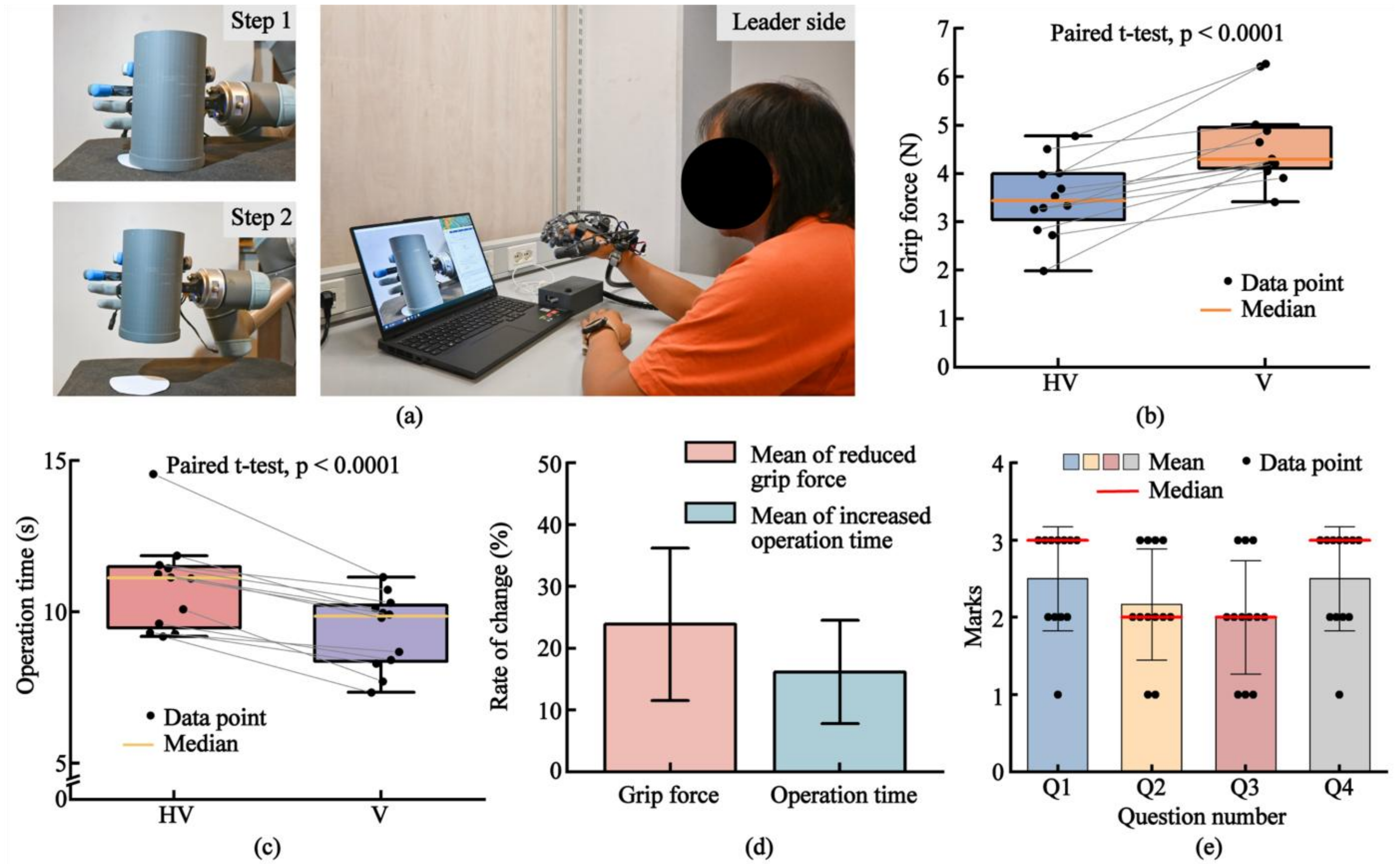

**Fig. 10 Experiment on teleoperated grasp task.** (a) Overview of the experimental setup. (b) Paired boxplot of grip force measured at the robotic fingertip under two feedback conditions. Paired t-test, $p<0.0001$. (c) Paired boxplot of operation time under the two feedback conditions. Paired t-test, $p<0.0001$. (d) Percentage changes in grip force and operation time across conditions. (e) Questionnaire results reported by participants after completing the trials. In (d) and (e), bars represent mean values and error bars indicate ±1 SD. Sample size: n = 12.

Condition V ($t(11) = 6.04$, $p < 0.0001$, Cohen's $d = 1.74$), with Condition V (mean = 4.614 N, SD = 0.871) showing significantly higher grip force than Condition HV (mean = 3.495 N, SD = 0.776). The mean difference was 1.119 N (95% CI: [0.712, 1.526] N). In the HV condition the robotic hand applied significantly less grip force. This reduction is indicative of the participants' capability to modulate the grasping force with the guidance of kinesthetic feedback, which is important in teleoperation applications for avoiding excessive force that could damage fragile objects or the robotic hand itself.

Since kinesthetic feedback involves demagnetization processes, additional time may be required during operation. Therefore, the task duration was also recorded. As depicted in Fig. 10(c), a paired-sample t-test revealed a statistically significant difference between Condition HV and Condition V ($t(11) = 6.63$, $p < 0.0001$, Cohen's $d = 1.91$), with Condition HV (mean = 10.87 s, SD = 1.53) showing significantly longer operation time than Condition V (mean = 9.37 s, SD = 1.23). The mean difference was 1.49 s (95% CI: [0.99, 1.99] s). As shown in Fig. 10(d), the average operation time increased by 16.2% with combined feedback. This increase was attributed to the processes of perceiving the kinesthetic feedback, deciding the level of force to apply, and transmitting the kinesthetic response. The mean grip force decreased by 23.9% under the same condition. This reduction could be further improved by enhancing the mapping accuracy between the human and robotic hands, which remains a key direction for future research.

Upon completing all trials, participants answered a questionnaire consisting of the following items:

Q1: How much did you rely on the kinesthetic feedback?

Q2: How much did you rely on the visual feedback?

Q3: How much was the kinesthetic feedback realistic (e.g. the feeling of the object)?

Q4: How much did kinesthetic feedback contribute to your confidence that the object was securely grasped during teleoperation?

Participants rated each item using a visual analog scale from −3 to +3. The results are summarized in Fig. 10(e). Overall, kinesthetic feedback reliance (Q1) and its contribution to grasp confidence (Q4) received the highest median scores, suggesting that participants actively engaged with the kinesthetic cues and found them valuable for maintaining stable grasps. In contrast, reliance on visual feedback (Q2) was slightly lower, and the realism of kinesthetic feedback (Q3) scored moderately—indicating potential for future improvement in feedback fidelity.

## VI. CONCLUSION

This paper presents a hand exoskeleton system utilizing MR brakes to provide kinesthetic feedback during teleoperated grasping. The proposed device exhibits a wide and tunable force feedback with negligible back-driving resistance, attributed to the field-dependent yield stress of the MRG and its interaction with the bearing structure. A kinematic model of the exoskeleton and a model of the MRG bearing were developed. For the MRG bearing model, the theoretical braking torque showed good agreement with experimental results. In practice, however, the magnetic field distribution and MRG filling conditions affect the overall performance, leading to discrepancies between theoretical and measured values. For the kinematic model, the theoretical force feedbacks exhibited less than 5 % relative error when the coil current ranged from 0.5 A to 1.25 A. These errors mainly stem from torque-measurement uncertainty, structural friction, and linkage-angle measurement inaccuracies. To evaluate usability, two user studies were conducted: an object-recognition task and a teleoperation task. The recognition task confirmed that the system enables accurate classification of objects with distinct stiffness levels, while accuracy decreased when distinguishing objects with similar stiffness. In the teleoperation experiment, users applied on average 28.2 % less grip force when kinesthetic feedback was provided, although task completion time increased by 16.5 %. Questionnaire results showed higher ratings for feedback reliance and grasp confidence, indicating the perceived usefulness of kinesthetic feedback. However, the realism of the feedback received moderate scores, likely due to the absence of other haptic modalities.

The proposed MR brake features a compact, lightweight, and high-torque design well suited for kinesthetic-feedback applications. Its wide torque range enables high force rendering while maintaining good backdrivability. Compared with motor-driven approaches, the proposed design offers significantly improved stiffness-rendering capability in a more compact form factor, enhancing the fidelity of kinesthetic rendering. Relative to passive devices reported in [26]–[30], it provides a wider feedback-force range and a more wearable dorsal-hand configuration that supports feedback for all fingers. The extended feedback range broadens the contact-force matching bandwidth in teleoperation, improving perceived realism and enabling safer manipulation of fragile or deformable objects. Therefore, the system shows strong potential for virtual-reality applications requiring immersive and stable force feedback.

## References and Footnotes

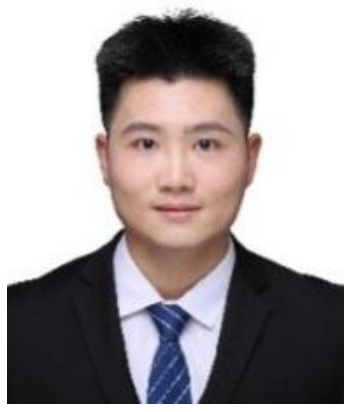

**Xianlong Mai** received the B.Sc. degree in theoretical and applied mechanics from University of Science and Technology of China in 2021, and now he is pursuing for Ph.D. degree in Instruments Science and Technology at University of Science and Technology of China. His research interests include wearable exoskeleton, human-robot interaction and teleoperation.

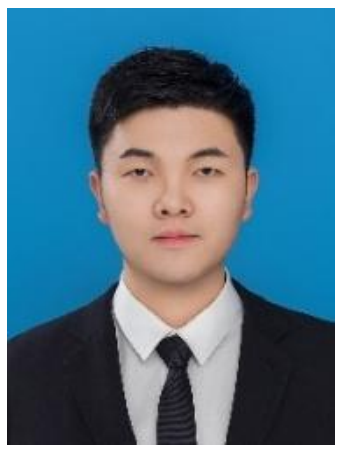

**Rui Yu** received the B.Sc. degree in automation from Anhui University in 2024, and now he is pursuing for an M.Sc. degree in Control Science and Engineering at Anhui University. His research interests include robotic arm control and wireless power transfer control.

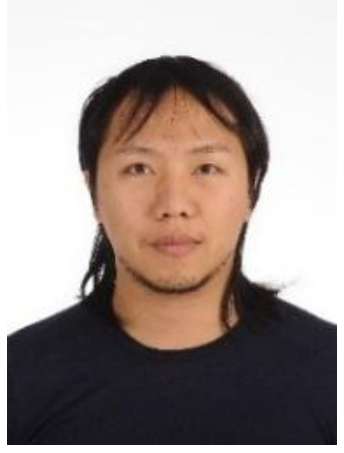

**Rui Chen** received the B.S. degree in mechanical engineering from Southwest Jiaotong University in 2019, dual M.S. degrees in product system engineering from RWTH Aachen University and in mechanical engineering from Tsinghua University in 2023. He is currently pursuing the Ph.D. degree in emerging digital technology at Sant'Anna School of Advanced Studies (SSSA), Pisa, Italy. His research interests include inflatable actuators and soft wearable robots.

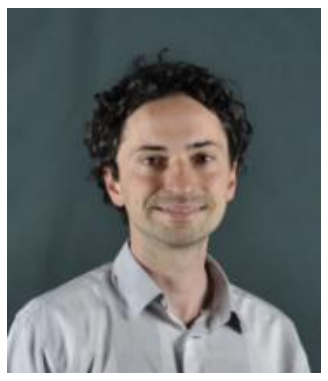

**Daniele Leonardis** is an Associate Professor of Mechanical Engineering at Scuola Superiore Sant'Anna, where he also earned his Ph.D. in Perceptual Robotics (2015). He holds an M.Sc. in Automation Engineering from the Polytechnic University of Bari (2009). His research focuses on haptic feedback, telemanipulation, and wearable robotics for clinical applications, and he leads national and European projects on fine haptic feedback in immersive VR for rehabilitation and teleoperation.

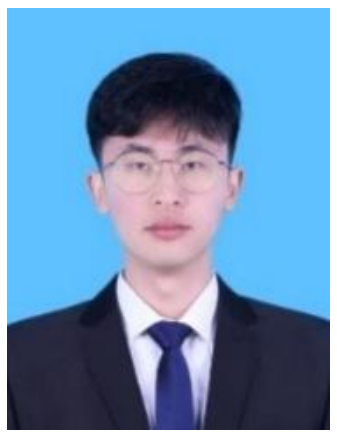

**Wenbo Li** received his B.E. degree in Naval Architecture and Marine Engineering from Northwestern Polytechnical University, He is currently pursuing the Master's degree with the School of Engineering Sciences at the University of Science and Technology of China. His research interests include magnetorheological clutch, safe actuator, exoskeleton, physical human-robot interaction safety, and wearable device.

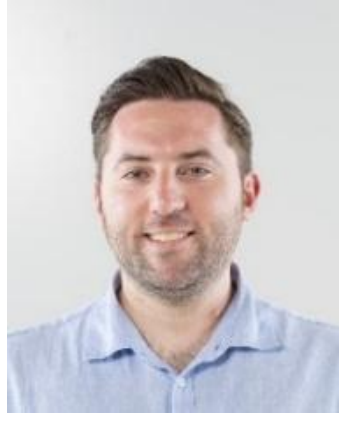

**Domenico Chiaradia** received his Ph.D. in perceptual robotics from Sant'Anna School of Advanced Studies (SSSA) in 2018. He is currently an Assistant Professor of mechanical engineering at SSSA, leading research on flexible and portable exoskeletons and soft exosuits for human–robot interaction (HRI). His research interests include physical HRI, rigid/flexible exoskeletons, soft exosuits for rehabilitation, flexible joint control, and haptic interfaces.

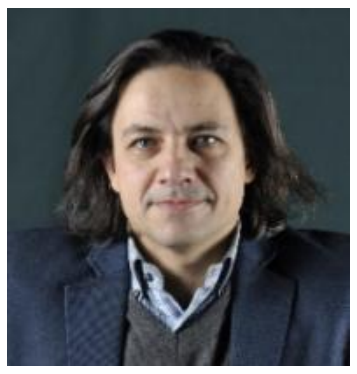

**Antonio Frisoli** (Senior Member, IEEE) is a Full Professor of engineering mechanics and robotics at Scuola Superiore Sant'Anna (SSSA), Italy, where he leads the Human-Robot Interaction and Intelligent Mechanics Institute. He co-founded two spin-offs (Wearable Robotics srl and Next Generation Robotics srl) and is active in technology transfer and open innovation. His research interests include collaborative, rehabilitation, and wearable robotics, exoskeletons, haptics, multimodal interaction, wearable devices, and virtual reality. He serves as an Associate Editor for IEEE Robotics & Automation Magazine and holds editorial roles in Eurohaptics, ICORR, and other major conferences and journals.

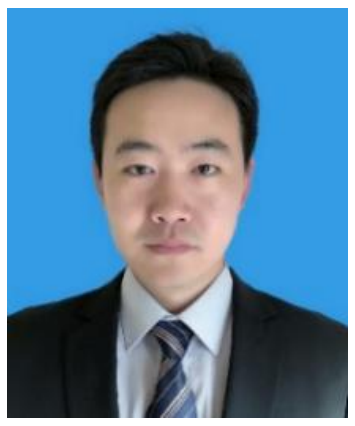

**Shuaishuai Sun** received the B.E. degree in mechanical engineering and automation from the China University of Mining and Technology, Beijing, China, in 2011, and the Ph.D. degree from the University of Wollongong, Wollongong, NSW, Australia, in 2016. His research interests include intelligent mechanical and mechatronics systems, system dynamics, smart materials and structures, innovative actuators for locomotive robots, and vibration control. He has published more than 120 journal articles in his research He is currently working as a Professor with the Department of Precision Machinery and Precision Instrumentation, University of Science and Technology of China, Hefei, China.